\documentclass[runningheads]{llncs}

 \usepackage[year=2024,ID=4247]{eccv}
 
\usepackage{eccvabbrv}

\usepackage{graphicx}
\usepackage{booktabs}
\usepackage{multirow}

\usepackage{booktabs} 

\usepackage[accsupp]{axessibility}  
\usepackage{xcolor}
\usepackage{relsize}
\usepackage{siunitx}
\usepackage{amssymb}
\usepackage{pifont}

\usepackage[pagebackref,breaklinks,colorlinks,citecolor=eccvblue]{hyperref}

\usepackage{orcidlink}

\begin{document}
 
\title{QuantTune: Optimizing Model Quantization with Adaptive Outlier-Driven Fine Tuning} 

\titlerunning{Abbreviated paper title}


\author{
Jiun-Man Chen\inst{1}\protect\footnotemark[1] and Yu-Hsuan Chao\inst{1}\thanks{These authors contributed equally to this paper.}  and Yu-Jie Wang\inst{1} and
Ming-Der Shieh\inst{1} and Chih-Chung Hsu\inst{1} and Wei-Fen Lin\inst{2}\thanks{This work was done independently of Rivos Inc.}}



\institute{National Cheng Kung University, Taiwan \and
Rivos Inc. USA}

\maketitle
\begin{abstract}
Transformer-based models have gained widespread popularity in both the computer vision (CV) and natural language processing (NLP) fields. However, significant challenges arise during post-training linear quantization, leading to noticeable reductions in inference accuracy. Our study focuses on uncovering the underlying causes of these accuracy drops and proposing a quantization-friendly fine-tuning method, \textbf{QuantTune}. Firstly, our analysis revealed that, on average, 65\% of quantization errors result from the precision loss incurred by the dynamic range amplification effect of outliers across the target Transformer-based models. Secondly, \textbf{QuantTune} adjusts weights based on the deviation of outlier activations and effectively constrains the dynamic ranges of the problematic activations. As a result, it successfully mitigates the negative impact of outliers on the inference accuracy of quantized models. Lastly, \textbf{QuantTune} can be seamlessly integrated into the back-propagation pass in the fine-tuning process without requiring extra complexity in inference software and hardware design. Our approach showcases significant improvements in post-training quantization across a range of Transformer-based models, including ViT, Bert-base, and OPT. QuantTune reduces accuracy drops by 12.09\% at 8-bit quantization and 33.8\% at 7-bit compared to top calibration methods, outperforming state-of-the-art solutions by over 18.84\% across ViT models.  

\end{abstract}

\section{Introduction}
\label{sec:intro}

Transformer-based models, including Vision Transformers (ViT) and BERT, have significantly advanced the field of machine learning by setting new performance benchmarks\cite{Attention2017, OPT2022, Llama2023, Bloom2022, Vit2020}. However, their evolution has led to a substantial increase in model complexity, characterized by an exponential rise in the number of parameters\cite{ComputeTrends2022, LargeLanguageModels2022}. This complexity introduces significant computational demands, resulting in considerable memory footprints, elevated power consumption, and increased inference latency. Such requirements pose substantial deployment challenges, especially on resource-constrained platforms such as mobile and IoT devices\cite{Hou2022MDVTC, Tang2022PatchSE, Ganesh2021BERTCompression, EdgeBERT2021, Liu2020FastBERT}.

\begin{figure*}[t]
  \centering
  \includegraphics[width=0.99\textwidth]{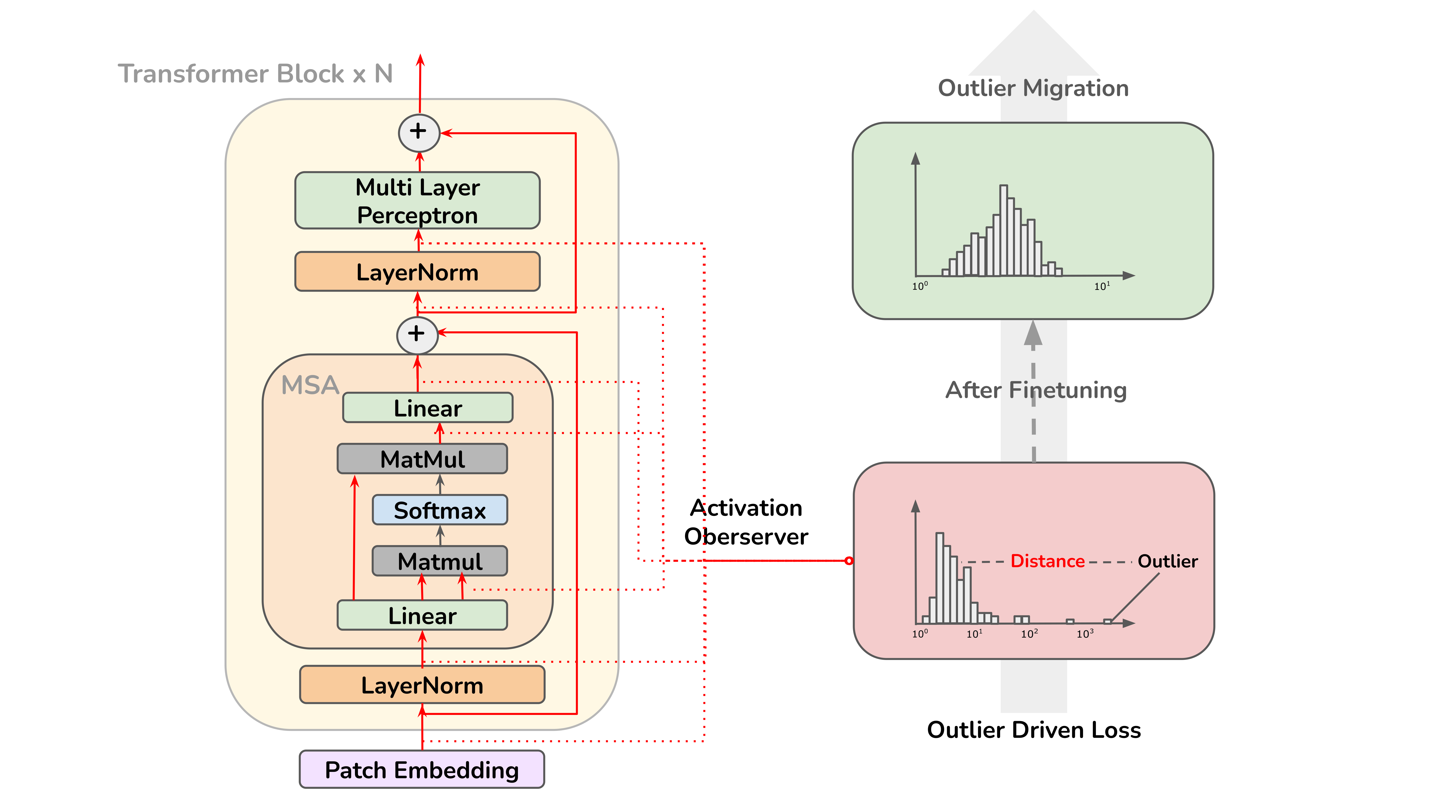}
\caption{Flowchart of the proposed QuantTune method, highlighting the use of the activation Observer to compute the outlier-driven loss, which mitigates outliers and reduces the dynamic range. The red line indicates the insertion point of the outlier observer.}
  \label{fig:flow}
  \vspace{-6.5mm}
\end{figure*}

Quantization emerges as an essential strategy for model compression, 
aiming to address these challenges by reducing model size and computational demands. 
Post-training dynamic range quantization is widely adopted, and it incurs substantial accuracy losses in many cases, especially drawing attention to Transformer-based models. A couple of prior research \cite{Bondarenko2021TransformerQuantization, SmoothQuant_2022, LLM_int8_2022} have mentioned that activation outliers could be the key contributing elements to these losses and propose different methods to alleviate the problem.  Our approach, depicted in Figure \ref{fig:flow}, employs the Activation Observer to calculate the outlier-driven loss, thereby mitigating the effects of outliers and potentially addressing one of the critical challenges observed in current quantization practices.

\begin{figure*}
  \centering
  \includegraphics[width=0.99\textwidth]{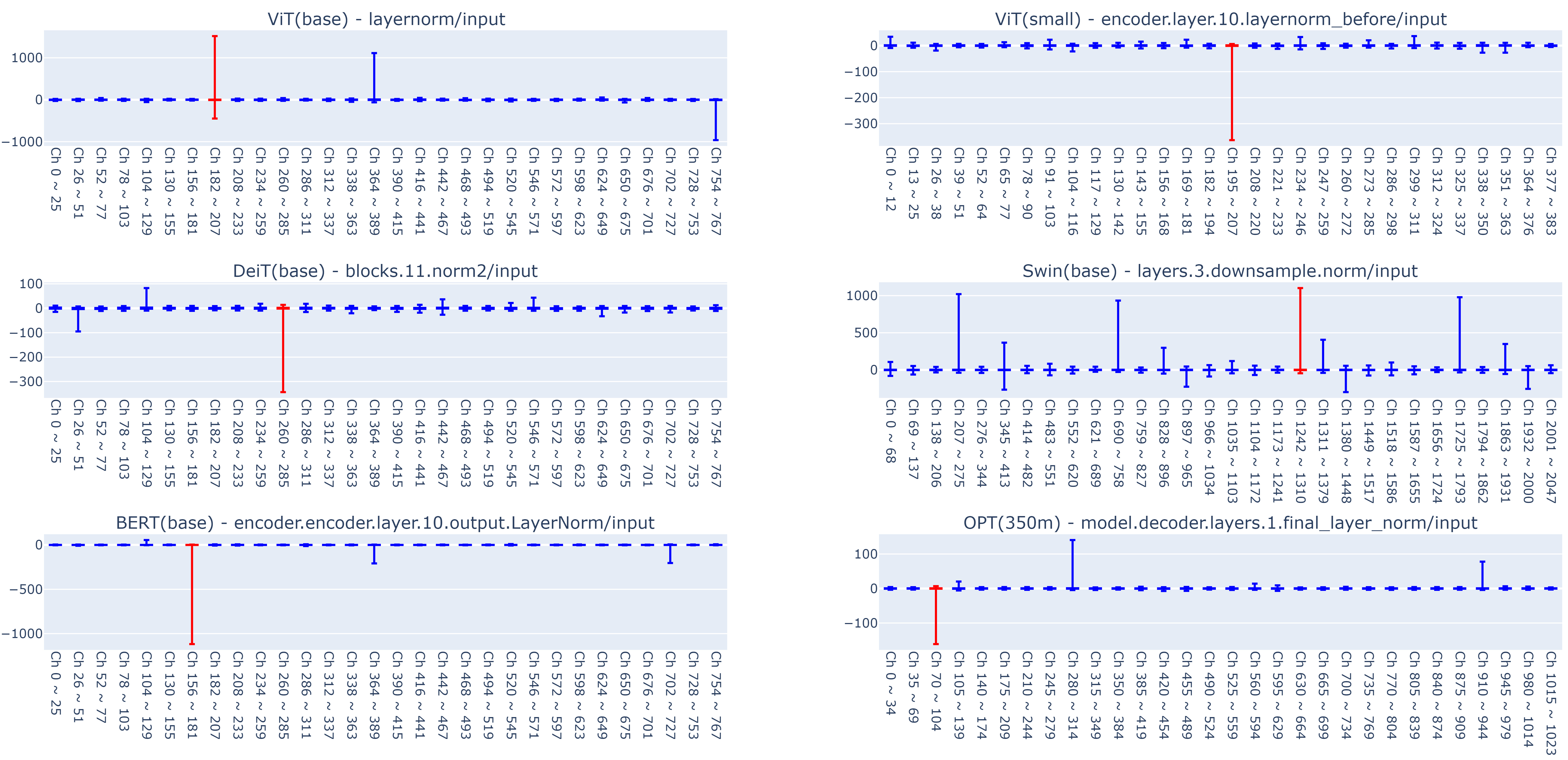}
  \caption{Comparative analysis of activation distributions across different Transformer models.
  Boxplots show activation value ranges in grouped channels for ViT (base), ViT (small), DeiT (base), Swin (base), BERT (base), and OPT (350m). Color denotes the activation value range, with red indicating the widest range. Data was segmented into 30-group segments for consistent comparison.}
  \label{fig:outlier}
  \vspace{-3mm}
\end{figure*}
Building on this premise, our analysis, illustrated in Figure \ref{fig:outlier}, confirms the presence of channel-wise outliers in Transformer-based models, including ViT \cite{Vit2020}, DeiT \cite{deit}, Swin \cite{Liu2021SwinTH}, BERT, and OPT. However, prior research does not fully elucidate why these outliers lead to a reduction in inference accuracy following post-training quantization (PTQ). It is speculated that the rounding errors from outliers directly contribute to a significant portion of the total quantization errors. Alternatively, outliers may indirectly cause an expansion in the dynamic range of activations, thereby significantly increasing the precision loss for non-outliers. Identifying the fundamental issue is crucial for determining if current state-of-the-art methods can be further improved or if alternative solutions are needed to address these issues. This paper initially focuses on uncovering these fundamental issues and evaluating contemporary approaches to establish our experimental baseline for subsequent optimization.

In our detailed analysis of the fundamental issue in Section III, we find that the primary concern is the expansive dynamic range caused by outliers. We recommend adopting a partial dynamic range for PTQ to counteract this. Traditional calibration methods are typically labor-intensive, time-consuming, and heavily reliant on specific datasets to determine optimal threshold settings, often failing to achieve the performance of the W32A32 baseline (where "W" represents the bit-width for weights, and "A" signifies the bit-width for activations), particularly in models like ViT-Base and ViT-Large. In contrast, our proposed method, termed \textbf{QuantTune}, utilizes outlier-driven techniques to manage the dynamic range expansion during fine-tuning, thus improving quantization accuracy and making Transformer-based models more amenable to quantization.

Our work broadens the scope of model quantization by investigating the effects of transitioning from W8A8 to W6A6 low-bit quantization across a variety of models, including ViT, DeiT, Swin, BERT, and OPT. The main contribution of this paper is threefold:
\begin{enumerate}
    \item \textbf{Model Adaptability:} QuantTune demonstrates robust adaptability across a wide range of Transformer architectures, effectively catering to both vision and language models. It is compatible with models having parameter counts ranging from 86 million to 350 million.
    \item \textbf{Low PTQ Performance Degradation:} Compared to the best calibration method, QuantTune decreases the average accuracy drop by 12.09\% at W8A8 quantization and surpasses the best calibration method by reducing accuracy loss by 33.8\% at W7A7. Furthermore, it outperforms state-of-the-art methods by reducing the accuracy drop by more than 18.84\% across all ViT models.
    \item \textbf{Hardware Independence:} QuantTune significantly reduces dependency on specific hardware toolchains for calibration, facilitating a quantization optimization process that is more accessible to software developers without specialized hardware. Moreover, it promotes uniform quantization, ensuring seamless compatibility with conventional computing platforms, including CPUs and GPUs.
\end{enumerate}

\section{Related Work}
Quantization is a technique that reduces computational time and memory usage in neural networks by employing low-bit representations for weights and activations \cite{Cheng2017SurveyMCADNN, Jacob2017Quantization, Nagel2019}. It is especially effective when using low-bit fixed-point formats, such as INT8, which offer improved energy efficiency over floating-point operations. According to \cite{Zhu2023SurveyMCLLM, Nagel2021}, quantization techniques are broadly categorized into two main approaches: QAT \cite{Choi2018PACTPC, Gholami2021, Wang2019HAQHA, Zhou2016DoReFANetTL, Esser2019LearnedSS} and PTQ \cite{Liu2021PostTrainingQF, Fang2020PostTrainingQuant, Krishnamoorthi2018QuantizingDCN}. While QAT can encounter scalability issues with large models, PTQ is deemed more suitable due to its training-free approach, conserving resources and enabling faster deployment without the need for access to the full dataset.

PTQ for Transformer-based models, including ViT \cite{Li2022RepQViTSP, Lin2022FQViTPQ} and large language models such as BERT \cite{Devlin2018BERTPO} and OPT \cite{OPT2022}, presents significant challenges in managing channel-wise outliers in activations during quantization \cite{SmoothQuant_2022, BERT_Busters_2021}. Notable discrepancies, often exceeding a thousandfold, in activation ranges across different channels can lead to substantial accuracy drops when employing per-tensor quantization. Studies have shown that these outliers frequently occur in the residual segments of Feed Forward layers \cite{Bondarenko2021TransformerQuantization, Quantizable_Transformers_2023}. Moreover, variations in the softmax and Multi-head self-attention mechanisms \cite{Outlier_Suppression_Plus_2023, Choukroun2022LowbitQO, Li2022QViTAF} further impact the accuracy of PTQ models.

Several methods employing non-uniform quantization have been developed to address the pronounced inter-channel variation in Vision Transformers. Lin et al. \cite{Lin2022FQViTPQ} introduced the Power-of-Two Factor (PTF), RepQ-ViT \cite{Li2022RepQViTSP} utilized scale reparameterization, and PTQ4ViT \cite{Yuan2022PTQ4ViTPF} developed Twin Uniform Quantization to mitigate asymmetric activations. In the realm of calibration optimization, OMSE \cite{Choukroun2022LowbitQO} focuses on minimizing the mean squared error, while APQ-ViT \cite{Ding2022TowardsAP} proposed block-wise strategies. Additionally, Q-ViT \cite{Li2022QViTAF} employed Distribution Guided Distillation for training-based improvements. Finally, PSAQ-ViT \cite{Li2022PSAQViTV2} introduced innovative PTQ methods targeting data-free applications.

Outlier generation in language models, often due to 'no-op' outcomes in attention mechanisms \cite{BERTology_2020}, is mitigated by various strategies. Reducing weight bit-width requirements \cite{AWQ_LLM_2023, LUT_GEMM_2022, GPTQ_2023}, refining quantization granularity \cite{Outlier_Suppression_2023}, and employing mixed-precision techniques in key areas \cite{LLM_int8_2022} 
have been explored. Additionally, new quantization combination algorithms for optimizing errors have been proposed \cite{LSQ_Plus_2020, Adaptive_Rounding_2020, ZeroQuant_V2_2023}. Recent works have introduced scaling and smoothing methods to adjust outliers pre-quantization, albeit increasing overhead \cite{SmoothQuant_2022, Outlier_Suppression_Plus_2023}. Finally, a novel approach with Gated Attention aims to address outlier generation fundamentally \cite{Quantizable_Transformers_2023}, but requires retraining and faces accuracy challenges in larger models like OPT-1.3B.

Most approaches address the activation outliers using diverse quantization strategies and calibration methods, typically involving non-uniform quantization like logarithmic scaling or specific calibration losses, sometimes necessitating specialized hardware for optimal execution. Different model components, such as linear layers, softmax, and layer normalization, often require distinct quantization approaches. Yet, these methods generally avoid using simpler, uniform, and symmetric quantization for the whole model, mainly due to inadequate handling of dynamic range issues. 
In contrast, based on our analysis detailed in Section III, we advocate utilizing a partial rather than full dynamic range for PTQ to manage activation outliers better. Our proposed method focuses on reducing the dynamic range, thereby enabling a more straightforward uniform and symmetric quantization approach across the entire model. This strategy aims to simplify the quantization process while preserving model performance. For further details, please refer to Section IV.

\begin{figure}[t]
  \vspace{0.5cm} 
  
  \begin{minipage}[b]{\textwidth}
    \centering
    \includegraphics[width=0.95\textwidth]{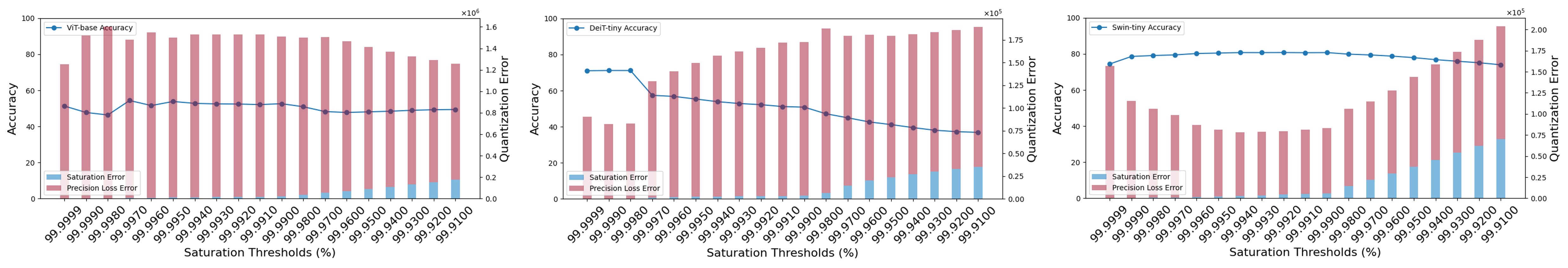}
    \caption{Accuracies and errors vary with different saturation thresholds across various ViT-relative models (left: ViT-base, middle: DeiT-tiny, right: Swin-tiny). The line chart displays the accuracies of ImageNet-1K corresponding to different saturation percentages. The bar chart illustrates two forms of error resulting from quantization: saturation error (blue bar) and precision loss error (red bar).}
    \label{Saturation_and_Precision}
  \end{minipage}
  \vspace{-6mm}
\end{figure}

\section{Fundamental Analysis for Quantization}

This section focuses on the essential task of identifying the fundamental causes of quantization errors and the limitations imposed by dynamic ranges in Transformer-based models. Understanding these fundamental issues is essential for designing a quantization-friendly learning mechanism, i.e., QuantTune. By pinpointing how outliers and dynamic range variations impact model accuracy and performance, we lay the groundwork for developing effective strategies that mitigate these effects, thereby enabling more efficient and accurate quantization processes. The emphasis on uncovering these underlying causes is pivotal for the subsequent introduction of QuantTune, which is aimed at enhancing model quantization without the need for complex hardware or extensive calibration efforts.

\subsection{Quantization Error Analysis}

\subsubsection{Quantization Error and Accuracy} This subsection delves into the nuanced relationship between outliers and model accuracy, building on insights from earlier research on full-precision models like those documented in \cite{BERT_Busters_2021}. While these initial studies underscored the critical role of outlier removal in affecting accuracy, they left the specific impact of outliers within quantized models largely uncharted.

Our investigation introduces a novel approach by employing an end-to-end search technique to ascertain the efficacy of using a partial, rather than a full, dynamic range for quantization. This strategy aims to saturate outliers within a specified limit. It establishes a saturation threshold, representing the percentage of the dynamic range remaining untouched, while values falling outside this threshold will be saturated. Figure \ref{Saturation_and_Precision} highlights our findings, presenting the optimal saturation thresholds necessary for maximizing accuracy across various models, including 99.999\% for ViT-base and DeiT-tiny, and 99.994\% for Swin-tiny. This analysis, which extends beyond the scope of previous studies such as \cite{BERT_Busters_2021}, illustrates the positive impact of controlling outliers on the performance of quantized models, thereby motivating us to develop the quantization-friendly learning framework.


\subsubsection{Saturation and Precision Loss} Utilizing partial dynamic range for quantization introduces two distinct forms of error, i.e., saturation and precision loss errors. Saturation error occurs when tensors are constrained to a fixed range, while precision loss error arises from scaling and rounding. Figure \ref{Saturation_and_Precision} reveals the relationship between two errors and shows that the precision loss error dominates the total error, accounting for at least 65\%. This finding further inspires us to design a precision loss-aware approach to ensure quantization-friendly capability. 


\begin{figure*}[t]
  \centering
  \includegraphics[width=0.99\textwidth]{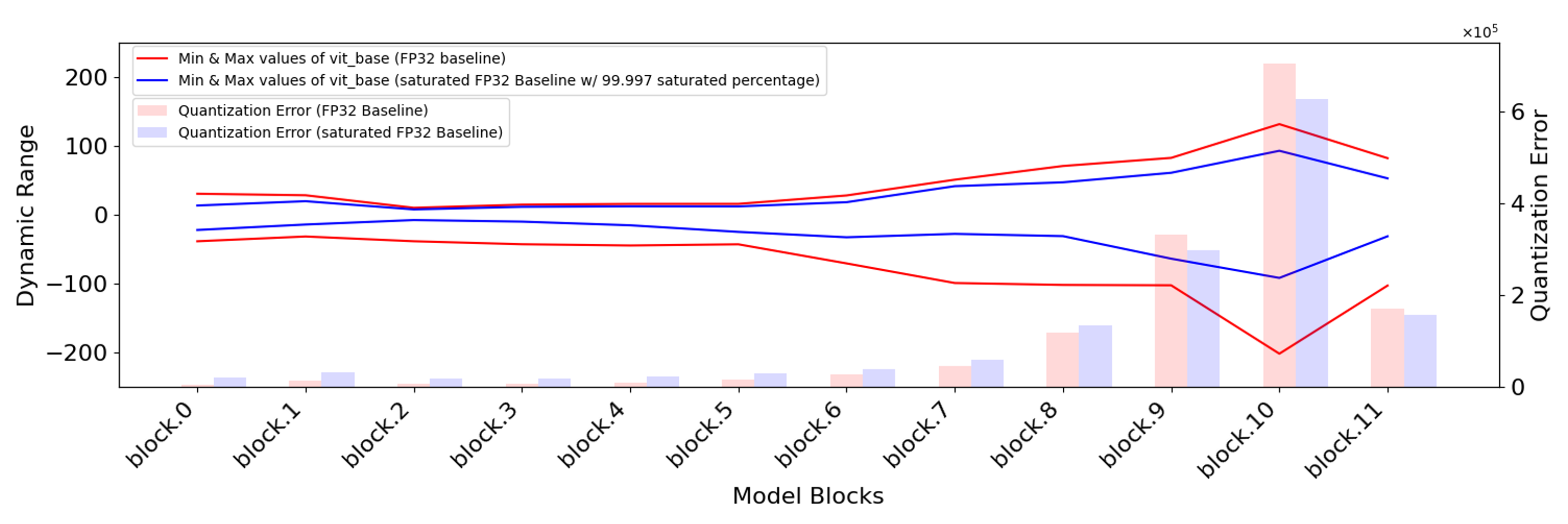}
  \caption{Precision loss error and dynamic range of each block in the ViT-Base model. The red line chart shows the dynamic range before saturation, while the blue line illustrates the dynamic range after saturation. The bar chart demonstrates the relative precision loss error, which equals the sum of the KL-divergence between full-precision and quantized tensors.}
  \label{Precision_Loss_Error_per_operator}
  \vspace{-2.5mm}
\end{figure*}

\subsection{Dynamic Range and Limitations}
\subsubsection{Precision Loss and Dynamic Range} Dynamic range is essential for precision loss error. In Figure \ref{Precision_Loss_Error_per_operator}, we compare the differences in precision loss error resulting from using full and saturated activations. As intuitively anticipated, quantization with saturated activations yields smaller errors, effectively reducing the dynamic range and providing higher accuracy after quantization simultaneously. Consistent with previous works \cite{Bondarenko2021TransformerQuantization, Quantizable_Transformers_2023}, our analysis demonstrates that within Transformer-based models, the dynamic range of output tensors grows larger with deeper depth. This extensive dynamic range contributes to significant precision loss, even when tensors are saturated.

\begin{figure*}[t]
  \centering
  \includegraphics[width=0.99\textwidth, keepaspectratio]{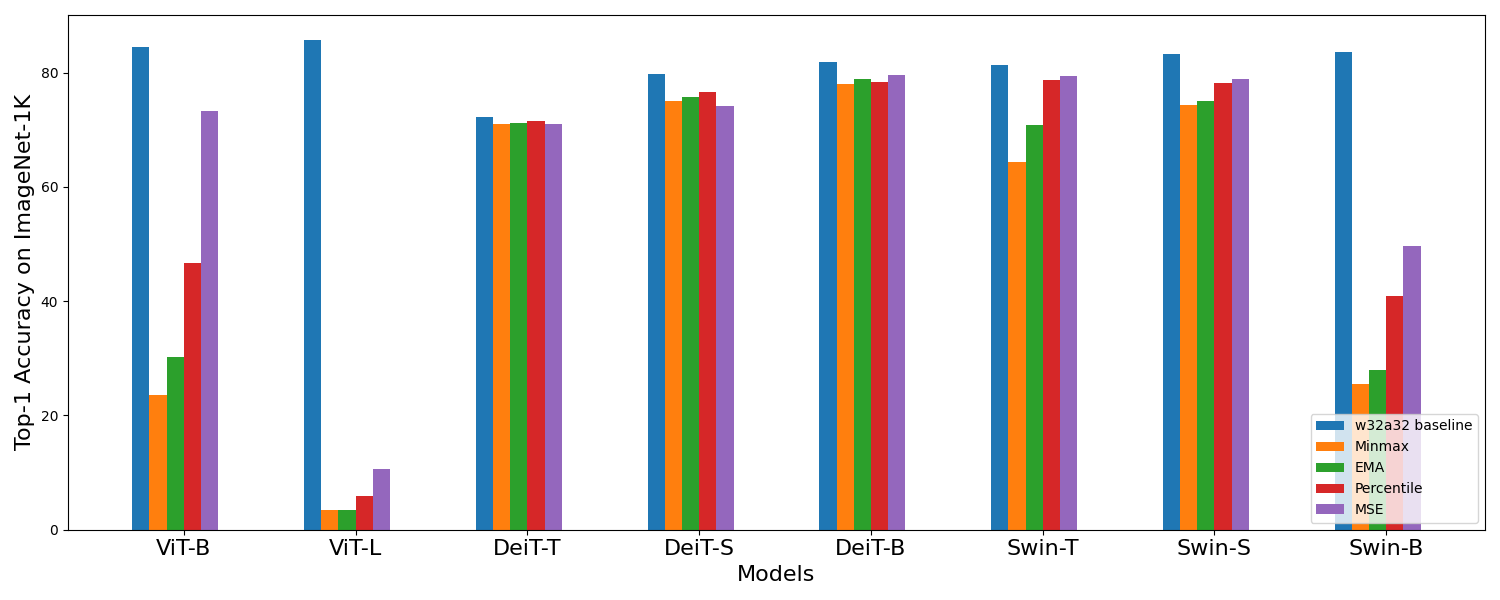}
  \caption{Performance of Transformer-based models with different calibration methods. This bar chart compares the top-1 accuracy on ImageNet-1K for various ViT architectures following different calibration methods.}
  \label{ViT_calibration_comparison}
  \vspace{-5mm}
\end{figure*}
Our finding aligns with previous studies; for instance, Jacob et al. \cite{Jacob2017Quantization} mentioned the negative effects of outliers in quantized models. Moreover, we find that the quantization errors of outliers are not large enough to dominate the inference accuracy drop. Instead, by broadening the dynamic range, outliers indirectly impose a more significant impact, making the overall data less precise and thus lowering the quantized model's performance. Our findings emphasize the critical importance of managing dynamic range to alleviate precision loss and, consequently, increase quantization efficacy. This serves as the primary motivation behind the development of our method, QuantTune, which is elaborated upon in Section 4.

\subsubsection{Saturation Impact on Model Performance} We adopt min-max, Mean Squared Error (MSE), Exponential Moving Average (EMA)\cite{Jacob2017Quantization}, and percentile-based approaches \cite{Li2019FullyQN} as the calibration methods for assessing the top-1 accuracy of ViT, Data-efficient Image Transformers (DeiT), and Swin-Transformers under ImageNet-1K dataset. Our findings, as shown in Figure \ref{ViT_calibration_comparison}, reveal that despite employing advanced calibration strategies at a W8A8 bit-width, there remains a noticeable performance gap compared to the W32A32 baseline, which is particularly pronounced in ViT-Base and ViT-Large models. These results suggest that even with careful calibration, achieving compatible results with baseline performance is challenging.


To accomplish this problem, some studies designed complicated observers or even sophisticated non-uniform quantizers for quantization; however, doing so introduces a significant time cost and is not always effective in closing the gap with the baseline. 

To address these challenges, QuantTune aims to eliminate the search overhead in the calibration process and provides a novel way to eliminate the impact of outliers in Transformer-based models to ensure a quantization-friendly architecture. 


\section{Proposed QuantTune}

 Building on the insights gained from our fundamental analysis for quantization, the challenge of precision errors in quantization, primarily due to rounding and scaling, prompts the need for a novel approach to judiciously adjusting the dynamic range of activations. Our QuantTune is thus specifically designed to mitigate the adverse effects associated with dynamic range constraints, thereby reducing precision loss after quantization.
 
 To address this challenge, a novel outlier-driven loss is proposed in this study to suppress activation outliers dynamically and judiciously by normalizing outlier effects, leading to more consistent activation patterns during the training phase, as drawn in Figure \ref{fig:flow}. So, our QuantTune is designed to strengthen the model's ability to withstand errors caused by quantization, highlighting our dedication to developing strategies that make quantization more effective. We draw the details of the proposed QuantTune in the following subsections.

\subsection{Proposed Outlier-Driven Loss}
This section will concentrate on how our novel outlier-driven loss is seamlessly integrated into the fine-tuning phase, marking a pivotal step in enhancing model resilience against quantization-induced errors without increasing training costs or requiring a long search time of calibration. 

To provide a solid foundation for our outlier-driven loss, it is essential to understand the standard loss functions typically employed in downstream tasks for models such as ViT and BERT. The most common loss function used in classification could be cross-entropy, as follows:

\begin{equation}
\ell_{\text{cls}} = -\frac{1}{m} \sum_{i=1}^{m} \sum_{j=1}^{k} y_{i,j} \log(\hat{y}_{i,j}),
\end{equation}
where \(m\) denotes the number of samples, \(k\) represents the number of classes, \(\hat{y}_{i,j}\) refers to the model's predicted probabilities for class \(j\) of the \(i^{th}\) sample, and \(y_{i,j}\) signifies the actual class of the \(i^{th}\) sample, typically expressed in a one-hot encoded vector, paralleling \(y_i\).

Our outlier-driven loss is designed to quantify and adjust for the divergence of activation from their expected statistical norms across the entire model by computing the normalized difference between the maximum absolute value of the activation tensor and its median across each layer in a batch.
During the forward pass, the observer calculates our outlier-driven loss function, targeting critical junctures within the model as highlighted by the red line in Figure \ref{fig:flow}. This observer is strategically placed to monitor both inputs and outputs of linear and LayerNorm layers, optimizing the dynamic range as follows:

\begin{equation}
    \ell_{\text{out}} = \frac{1}{m}  \frac{1}{n}  \sum_{i=1}^{m} \sum_{j=1}^{n} \left( \frac{\max(\lvert A_{j,i} \rvert) - \text{median}(\lvert A_{j,i} \rvert)}{\sigma(A_{j,i})} \right), 
\end{equation}
where $n$ denotes the specific instances where the loss is applied, starting from $j=1$, encompassing both the input and output tensors of linear layers, as well as those of LayerNorm layers within the architecture. $A_{j,i}$ signifies the activation tensor for the instance indexed by $j$ for the $i^{th}$ sample in the batch. The loss function calculates the normalized difference between the maximum absolute value $\max(\lvert A_{j,i} \rvert)$ and the median value $\text{median}(\lvert A_{j,i} \rvert)$, compared to the standard deviation $\sigma(A_{j,i})$. This ensures comprehensive coverage across both the individual samples and the targeted layers or layer aspects, enhancing the model's robustness by mitigating the impact of outliers.

To effectively benefit from the advantages of our outlier-driven loss, we judiciously integrate the proposed outlier-driven loss with the regularizer to balance the dynamic range during the training phase as follows:
\begin{equation}
  \begin{aligned}
  \ell_{\text{t}} &= (1 - \alpha) \cdot \ell_{\text{cls}} + \alpha \cdot \ell_{\text{out}},
  \end{aligned}
\end{equation}
where \(\ell_{\text{cls}}\) represents the conventional loss function used for the primary task, \(\ell_{\text{out}}\) denotes 
our specially designed outlier-driven loss, and \(\alpha\) serves as the balance factor between these two components. By tuning the balance factor 
\(\alpha\), which ranges between 0 and 1, is vital in our methodology as it moderates between standard and outlier-driven losses, optimizing the model's handling of outliers without forfeiting its main task efficiency.

\begin{figure*}[!htbp]
  \centering
  \includegraphics[width=0.99\textwidth]{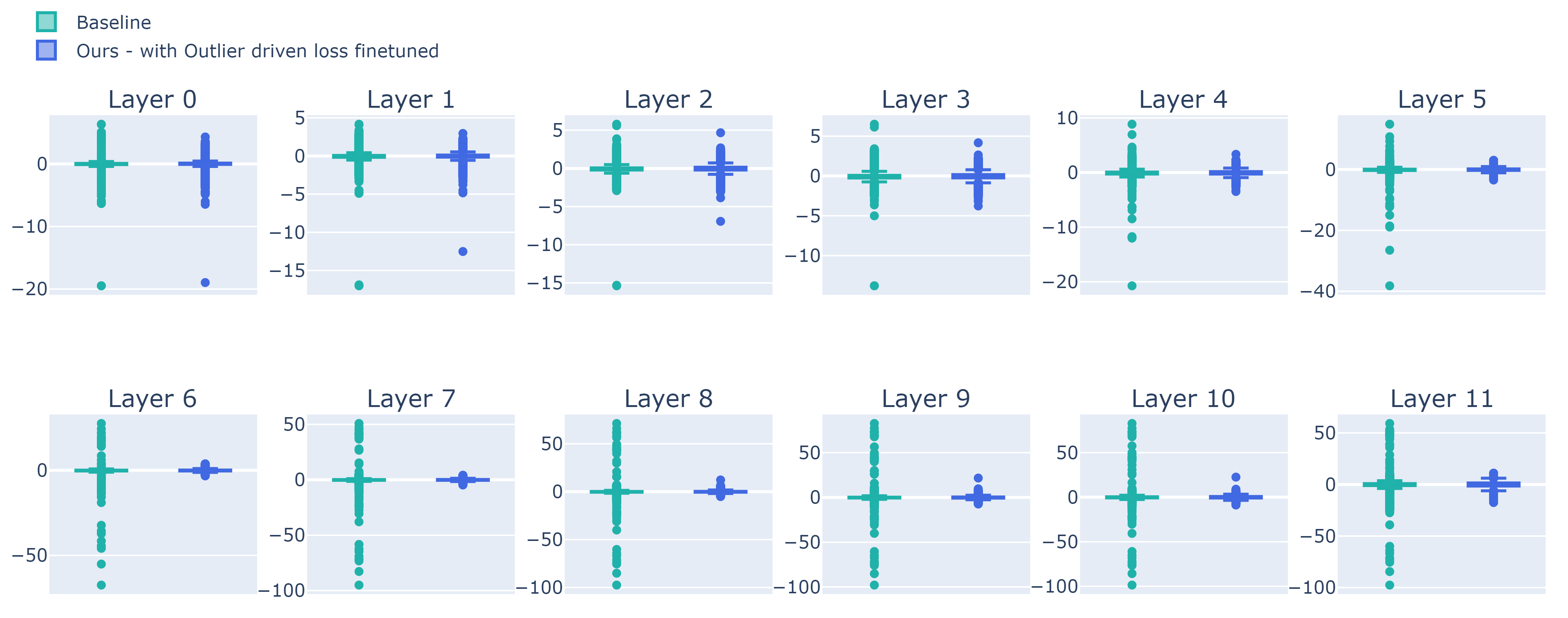}
  \caption{Dynamic range comparison of activation before the second LayerNorm layer within the Transformer blocks, illustrating differences between the baseline ViT-base model and the model fine-tuned with outlier-driven loss.}  
  \label{fig:activation_diff}
  \vspace{-3mm}
\end{figure*}

\subsection{Dynamic Range Optimization: The QuantTune Advantage}

We implement QuantTune, incorporating the outlier-driven loss to address the issue of outliers directly, thereby significantly reducing the dynamic range of activations within the model. This improvement is particularly noticeable in the inputs and outputs around linear and LayerNorm layers, areas previously identified as critical for dynamic range complications. Our tailored approach leads to a more uniform distribution of activation values, as evidenced by the reduced spread in dynamic range, effectively rendering the model more quantization-friendly. Moreover, the proposed QuantTune could be treated as a plug-and-play framework, which can seamlessly integrate with any existing quantization strategies, such as non-uniform quantization, to reduce performance degradation, making QuantTune more practice.

This reduction in dynamic range, as demonstrated in Figure \ref{fig:activation_diff}, not only aligns with previous findings\cite{Bondarenko2021TransformerQuantization, Quantizable_Transformers_2023} but also extends them by providing a practical solution to the identified issues. By constraining outlier activations, we directly decrease precision errors, which is the root cause of significant loss, thereby enhancing the model's overall accuracy after quantization. The resultant decrease in dynamic range across layers underscores the efficacy of the QuantTune method in creating models better suited for PTQ, marking a significant step forward in addressing the longstanding challenge of maintaining accuracy in quantized models.

\section{Experiments}
To validate the versatility of our approach across different Transformer-based models, this section delves into the comprehensive evaluation of our innovative outlier-driven technique on a variety of vision Transformers and language models.
In vision Transformer selection, we assess the performance of our proposed QuantTune based on ViT \cite{Vit2020}, DeiT \cite{deit}, and Swin-Transformer \cite{Liu2021SwinTH}. We utilized the top-1 accuracy metric on the ImageNet-1K \cite{Russakovsky2014ILSVRC, Deng2009ImagenetAL} validation set to assess the efficacy of the proposed methods.
During training, we allocated 10\% of the training data for validation purposes and subsequently evaluated the model performance using the entire validation dataset upon the completion of training.

As for language models, we utilized the BERT-base-uncased model, which features 109M parameters and has been pre-trained using the masked language modeling (MLM) strategy directly from HuggingFace's libraries for our fine-tuning purposes. Additionally, we assessed the 1.3B parameter variant of OPT, pre-trained with the causal language modeling (CLM) strategy. Due to computational constraints, we adapted our training to a maximum sequence length of 512. For evaluating the Bert model, we utilized the GLUE benchmark \cite{Wang2018GLUEMT}. Furthermore, to enhance our evaluation of OPT model capabilities, we assessed the performance of OPT models using the LAMBADA dataset \cite{Paperno2016TheLD}.

\subsection{Experiment Setup}
 \subsubsection{Quantization Scheme.} 
 
 To ensure a fair comparison, we applied identical quantization methods to all models compared in this study. Standard symmetric and uniform quantization for both activations and weights are used by the min-max strategy to determine the quantization ranges for our models. We apply whole-model quantization except for softmax and layer normalization layers for our QuantTune. This exclusion is because the computational demands of Transformer-based models do not primarily reside within these layers. 
 To facilitate a comparison with baseline models calibrated for quantization, we established fixed parameters for calibration, setting the number of batches to 10, with each batch containing 100 images. Our extensive analysis covered whole-model quantization levels ranging from W8A8 to W6A6. The goal was to identify an optimal balance between computational efficiency and the preservation of model integrity.
 
\subsubsection{Tuning of Outlier-Driven Loss Hyperparameter.}  

The \(\alpha\) parameter for our outlier-driven loss was fine-tuned within the range of 0 to 1, specifically at intervals [0.3, 0.5, 0.7], to find the optimal balance between outlier correction and maintaining performance. We implemented an \(\alpha\) decay strategy to gradually reduce its influence, allowing a seamless shift from focusing on outliers to prioritizing main task accuracy tailored to model needs.

\subsubsection{Comparison with State-of-the-Art Methods.} 

We evaluate the performance of selected Transformers based on the proposed QuantTune and other peer methods associated with the calibrated-based approach. This encompasses established methods such as min-max, Mean Squared Error (MSE), Exponential Moving Average (EMA)\cite{Jacob2017Quantization}, and percentile approaches as discussed in Li et al.\cite{Li2019FullyQN}, along with the Minimum MSE Quantization (OMSE) introduced by Choukroun et al.\cite{Choukroun2022LowbitQO}.

We examine the FQ-ViT approach by Lin et al.\cite{Lin2022FQViTPQ}, which utilizes the Power-of-Two Factor (PTF) technique. Additionally, we evaluate against PTQ4ViT by Yuan et al.\cite{Yuan2022PTQ4ViTPF}, which introduces Twin Uniform Quantization specifically designed for asymmetric distributions, complemented by a Hessian-guided metric for determining the optimal scaling factor. We explore a novel approach that utilizes Gated Attention to tackle outlier issues fundamentally \cite{Quantizable_Transformers_2023}. Through this comprehensive assessment, we aim to demonstrate the unique advantages and robustness of our QuantTune method against a backdrop of both conventional and modern quantization strategies.

\begin{table}[t]
\centering
\sisetup{detect-weight,mode=text}
\renewrobustcmd{\bfseries}{\fontseries{b}\selectfont}
\renewrobustcmd{\boldmath}{}
\newrobustcmd{\B}{\bfseries}
\small
\begin{tabular}{l c c c c c c c c c c c}
  \toprule
  \multirow{2}{*}{\B Method} & \multirow{1}{*}{\B W/A } & \multirow{2}{*}{\B Ave} & \multicolumn{3}{c}{\B ViT} & \multicolumn{3}{c}{\B DeiT} & \multicolumn{3}{c}{\B Swin} \\
  \cmidrule(lr){4-6} \cmidrule(lr){7-9} \cmidrule(lr){10-12}
   &\B Bit && \B Small & \B Base & \B Large & \B Tiny & \B Small & \B Base & \B Tiny & \B Small & \B Base \\
  \midrule
  Baseline & 32/32 &81.45& 80.57&	84.53	&85.81&72.21	&79.85&	81.85&81.38&	83.23&	83.60\\
  \midrule
  Minmax & 8/8 &32.77& 31.70	&3.37&2.08&	70.26&	56.34	&33.56&66.53&	22.34&	8.74\\
  EMA\cite{Jacob2017Quantization} & 8/8 &35.30& 37.61	& 3.6 &2.11&	70.16	&61.13	&40.10&69.14&	24.49	&9.35\\
  Percentile\cite{Li2019FullyQN} & 8/8 &47.10& 48.32	& 29.14&3.52&	70.67	&71.17	&72.10&75.55	&40.53	&12.89\\
  OMSE\cite{Choukroun2022LowbitQO} & 8/8 &65.33& 71.03&	74.26&11.43&	\B 71.06	&74.70	&\B 79.86&\B 79.85	&79.19	&46.57\\
  \midrule
  QuantTune (ours) & 8/8 &\B 77.42&  \B 76.23 &  \B 79.67 & \B 79.24 & 70.16 & \B 77.55 &  79.32 &77.93 &\B 79.23 & \B 77.46\\
  \midrule

  Minmax & 7/7 &7.30& 0.41	&0.17&	0.71&	61.09	&0.46&	0.43	&2.13&	0.15	&0.13\\
  EMA\cite{Jacob2017Quantization} & 7/7&7.67& 0.38 &	0.15&0.86 &63.45	 &0.56 &0.52 &2.80&	0.14&0.14\\
  Percentile\cite{Li2019FullyQN} & 7/7 &10.66& 0.45 & 	0.47 & 	1.32	 & 65.79	 & 9.34	 & 0.70	 & 17.45 & 	0.17 & 	0.21\\
  OMSE\cite{Choukroun2022LowbitQO} & 7/7 &34.95& 9.15&	11.25&	2.00	&\B 67.56	&51.94&	72.13	&52.43	&39.17&	8.93\\
  \midrule
  QuantTune (ours) & 7/7 &\B 68.75& \B  69.39 &	\B 50.94 &	\B 77.04	 &63.24	 &\B 64.50	 &\B 73.40 & \B 75.26 & \B 71.28 & \B 73.68\\
  \bottomrule
\end{tabular} 
\vspace{3mm} 
\caption{Comparison with different calibration methods for symmetric uniform quantization in ViT \cite{Vit2020}, DeiT \cite{deit}, and Swin-Transformer \cite{Liu2021SwinTH} evaluated on ImageNet-1K with top-1 accuracy.}
\label{tab:ViT_ImageNet}
\vspace{-5mm}
\end{table}
\raggedbottom

\subsection{Performance Evaluation}
As illustrated in Table \ref{tab:ViT_ImageNet}, we compared QuantTune against standard calibration methods, where the best-performing calibration method, OMSE, achieves an average accuracy of 65.33\% across these models. In contrast, QuantTune significantly diminishes the average accuracy drop from 16.12\% to merely 4.03\% at W8A8 quantization on the ImageNet-1K dataset. For W7A7 quantization, QuantTune further reduces the accuracy drop by over 33.8\% compared to OMSE, showing promising results of our QuantTune.

\begin{figure*}[!ht]
  \centering
  \includegraphics[width=0.99\textwidth]{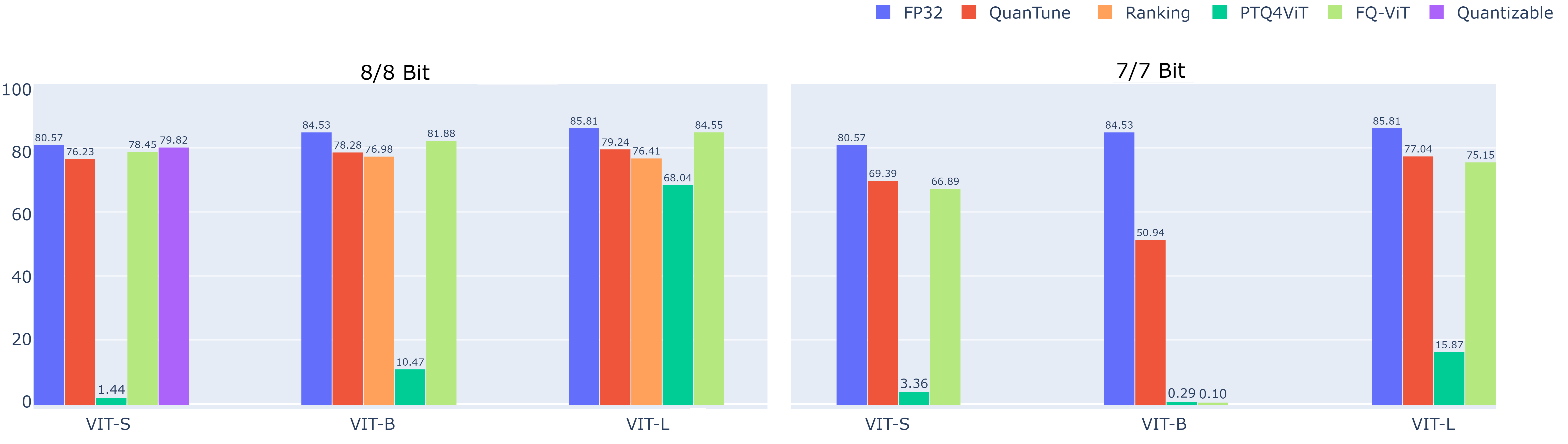}
  \caption{State-of-the-art ViT model comparison via ImageNet-1K Top-1 accuracy. Accuracy for Quantizable and Ranking was sourced directly from publications due to the unavailable code. PTQ4ViT and FQ-ViT performance could be altered by our stricter quantization approach versus the original methods.}  
  \label{fig:stoa_comparison}
\end{figure*}

Figure \ref{fig:stoa_comparison} shows the performance evaluation between our QuantTune and other peer methods. Our QuantTune, offers a comparable performance in top-1 accuracy metric, compared with leading methods like FQ-VIT\cite{Lin2022FQViTPQ}, Quantizable\cite{Quantizable_Transformers_2023}, and Ranking\cite{Liu2021PostTrainingQF} at 8-bit quantization, also significantly surpasses PTQ4ViT\cite{Yuan2022PTQ4ViTPF}, which shows minimal effectiveness at 1.44\% for ViT-S and 10.47\% for ViT-B in 8-bit quantization settings. Furthermore, we observe significant declines in performance among other methods; notably, FQ-VIT's accuracy plummets to just 0.1\% for the ViT-B model under lower-bit quantization (i.e., 7-bit). In contrast, QuantTune sustains a remarkable accuracy rate, reducing the accuracy drop by over 18.84\% compared to FQ-VIT, averaged across ViT-S, ViT-B, and ViT-L models. This showcases QuantTune's superior capability in low-bit scenarios. Such substantial improvement stems from QuantTune's adept management of dynamic ranges.

\begin{table}[!ht]
  
  \centering
  \scriptsize
  \sisetup{detect-weight,mode=text}
  \renewrobustcmd{\bfseries}{\fontseries{b}\selectfont}
  \renewrobustcmd{\boldmath}{}
  \newrobustcmd{\B}{\bfseries}
  \begin{minipage}{0.53\textwidth} 
  \begin{tabular}{lccc}
  \toprule
  \textbf{\multirow{3}{*}{Method}} & 
  \textbf{\multirow{3}{1.5cm}{\centering   Hardware friendly}} &
  \textbf{\multirow{3}{1.2cm}{\centering   No retrain}} & 
  \textbf{\multirow{3}{1.7cm}{\centering   No calibration search}}  \\
&&&     \\
&&&     \\
  \midrule
    PTQ4ViT\scriptsize{*}\cite{Yuan2022PTQ4ViTPF} &\checkmark &\checkmark & \texttimes  \\
    FQ-ViT\scriptsize{*}\cite{Lin2022FQViTPQ} &\texttimes&\checkmark&\texttimes  \\
    Quantizable\scriptsize{\dag}\cite{Quantizable_Transformers_2023}&\checkmark&\texttimes&\checkmark  \\
    Ranking\scriptsize{\dag} \cite{Liu2021PostTrainingQF}&\checkmark&\checkmark&\texttimes  \\
\midrule
    QuantTune (ours) &\checkmark&\checkmark&\checkmark  \\
\bottomrule
\vspace{2pt}
  \end{tabular}
  
 \caption{Overhead requirements and comparison between state-of-the-art quantization approaches and QuantTune.}
 \label{tab:ViT_ImageNet2}
\vspace{-1mm}
\end{minipage}
\hfill
\hspace{0.9mm}
\begin{minipage}{0.41\textwidth} 
\centering
    
    \sisetup{detect-weight,mode=text}
    \centering
      {\fontsize{8pt}{7pt}\selectfont
    \begin{tabular}{lccc}
      \toprule
       \B Method &  \B  W/A Bit  & \B OPT-350m  \\
        \midrule
      Baseline  & 32/32 & 67.57   \\
      \midrule
      Minmax    & 8/8   & 58.29   \\
      OMSE\cite{Choukroun2022LowbitQO}      & 8/8   & 9.57   \\
      Percentile\cite{Li2019FullyQN} & 8/8   & 8.85   \\
    \midrule
      QuantTune (ours)     & 8/8   & \B 62.50   \\
      \bottomrule
    \end{tabular}
    \vspace{6.5 pt}
    \caption{Quantization performance comparison with calibration methods for OPT Models on Lambada dataset.}
    \label{tab:OPT_LAMBADA}
      \par} 
      \end{minipage}
    \vspace{-5mm}
\end{table}

Considering the overhead requirements, we show that the proposed QuantTune could achieve the best trade-off between the performance and overhead, as shown in Table \ref{tab:ViT_ImageNet2}
For instance, PTQ4ViT and Ranking require specialized hardware support for calibration search, which can be both time-consuming and costly and could be useless when the test set is changed. In addition, FQ-ViT employs a non-uniform quantization method, necessitating specialized hardware for efficient processing. Furthermore, 
QuantTune eliminates does not require re-training, presenting a more time-efficient and feasible solution.
Conversely, QuantTune circumvents these issues by addressing outlier problems during the fine-tuning stage, thus eliminating the need for extensive retraining, calibration search, and reliance on specialized hardware. This not only makes QuantTune more feasible but also enables its plug-and-play compatibility with straightforward uniform and symmetric quantization approaches. This ensures seamless integration with standard computing platforms like CPUs and GPUs, further reducing the requirement for specialized hardware and making it a cost-effective solution for model quantization.

Regarding OPT models as detailed in Table \ref{tab:OPT_LAMBADA}, QuantTune reduces the accuracy drop by 4.24\% compared to the min-max calibration method. Moreover, for BERT as shown in Table \ref{tab:GLUE}, our method achieves no loss in accuracy at 8-bit quantization and reduces the accuracy drop by 5.95\% at 6-bit quantization. This demonstrates the effectiveness of our method even in low-bit scenarios and its applicability across various models.

\begin{table}[t]

    \centering
    \sisetup{detect-weight,mode=text}
    \renewrobustcmd{\bfseries}{\fontseries{b}\selectfont}
    \renewrobustcmd{\boldmath}{}
    \newrobustcmd{\B}{\bfseries}
  {\fontsize{8pt}{7pt}\selectfont
  \begin{tabular}{@{}lcccccccccc@{}}
    \toprule
     \B Method & \B W/A Bit & \B GLUE & \B CoLA  & \B SST2  & \B MRPC  & \B STS-B &  \B QQP  & \B MNLI{\scriptsize(m/mm)} & \B QNLI & \B RTE \\
    \midrule
    Baseline     & 32/32 & 81.25 & 55.41 & 89.45 & 86.52 & 89.15 & 90.64 & 81.14/81.42    & 90.02 & 67.51 \\ 
    \midrule
    OMSE\cite{Choukroun2022LowbitQO}       & 8/8   & 80.95 & 53.38 & 89.91 & 86.52 & 88.07 & 90.50 & 80.80/80.81    & 89.29 & 69.31 \\
    Minmax     & 8/8   & 80.37 & 52.62 & 88.65 & 85.05 & 87.94 & 90.42 & 80.77/80.94    & 89.07 & 67.87 \\
    Percentile\cite{Li2019FullyQN} & 8/8   & --    & --    & 91.74 & 85.78 & --    & 90.47 & 83.11/84.28    & 89.11 & 66.43 \\
\midrule    
    QuantTune (ours)       & 8/8   &  \B 81.54 & 53.03 & 92.66 & 86.52 & 87.89 & 90.42 & 83.06/83.95    & 89.58 & 66.79 \\
\bottomrule
\\[0.2pt]
    OMSE\cite{Choukroun2022LowbitQO}       & 6/6   & 53.69 & 18.89	& 84.29	& 32.35	& 56.58 &	75.59	& 48.25/48.28	   & 71.66 & 47.29 \\
    Minmax     & 6/6   & 47.87 & 6.26  & 80.62 & 67.40 & 42.50 & 65.90 & 34.98/35.30    & 50.54 & 47.29 \\
    Percentile\cite{Li2019FullyQN} & 6/6   & --    & --    & 72.25 & 32.11 & --    & 70.90 & 38.32/38.58    & 64.43 & 47.29 \\
    \midrule
    QuantTune (ours)      & 6/6   & \B 59.24 & 18.16 & 79.70 &   68.38 & 42.50 & 69.31 & 64.64/63.78    & 79.44 & 47.29 \\
    \bottomrule
  \end{tabular}
 \vspace{3mm} 
\caption{Performance comparison of QuantTune and calibration methods on BERT-Base model across GLUE benchmark tasks: STS-B and CoLA evaluated using Matthews correlation and Pearson correlation, respectively, with other tasks measured by accuracy, summarized by the GLUE average score.}  \label{tab:GLUE}
  \par} 
  \vspace{-7mm}
\end{table}

While observing the effectiveness of QuantTune in our experiments, we acknowledge the potential for further significant impacts. Although our current validation is comprehensive, future work will aim to extend our methodology to larger-scale models, such as OPT-175B or LLAMA-70B. This expansion will facilitate a broader validation, fully showcasing the capabilities and adaptability of our approach.


\section{Conclusion}
\label{sec:conclusion}
In conclusion, we have demonstrated that the degradation in performance of quantized Transformer models can primarily be attributed to the extended dynamic ranges introduced by outliers, which compromise data precision and quantization accuracy. To address this challenge, we introduced QuantTune, a novel fine-tuning methodology that utilizes an outlier-driven loss function to regulate activation dynamic ranges effectively. By adjusting weights to account for outlier deviations, our approach systematically narrows the dynamic ranges, significantly mitigating quantization errors and reducing the adverse effects of outliers. Our empirical results are compelling: QuantTune reduces the average accuracy drop by 12.09\% at 8-bit quantization compared to calibration methods. Moreover, QuantTune exhibits outstanding performance even in low-bit scenarios (e.g., 7-bit, 6-bit), surpassing the best calibration method by minimizing accuracy loss by 33.8\% at 7-bit. Additionally, it surpasses existing state-of-the-art methods by decreasing the accuracy drop by more than 18.84\% across all ViT models. Additionally, QuantTune broadens its applicability to diverse model architectures, such as BERT and OPT, ensuring effective quantization while preserving strong performance and accuracy even at 6-bit.

Beyond its performance merits, QuantTune exemplifies model universality and dataset insensitivity, ensuring its applicability across various Transformer models and datasets. QuantTune seamlessly integrates into fine-tuning, demanding no additional time or computational complexity during inference. It also guarantees hardware independence, circumventing the need for specialized hardware for calibration and ensuring seamless compatibility with standard computing platforms. This positions QuantTune as a pioneering software-based solution for those aiming to enhance quantization efficiency.

\subsection*{Acknowledgements}
We greatly appreciate and thank Dr. Gaun-Ming Su for his insightful feedback and discussion on this project.


\clearpage  

%
%
\bibliographystyle{splncs04}
\bibliography{main}
\end{document}